\documentclass[letterpaper, 10 pt, conference]{ieeeconf}  

\IEEEoverridecommandlockouts                              

\usepackage{booktabs,balance,subfigure}
\usepackage{graphicx}
\usepackage{multirow}
\usepackage[sort,compress]{cite}
\let\labelindent\relax
\usepackage{enumitem}
\usepackage{xcolor}
\usepackage{amsmath}
\usepackage{hyperref}
\hypersetup{
    colorlinks=true,
    urlcolor=black,
    linkcolor=black
    }
\setlist[itemize]{align=parleft,left=0pt..1em}
\newcommand{\sysname}{WAG}

\title{\LARGE \bf
City-wide Street-to-Satellite Image Geolocalization\\%
of a Mobile Ground Agent
}

\author{Lena M. Downes$^{1}$, Dong-Ki Kim$^{2}$, Ted J. Steiner$^{3}$ and Jonathan P. How$^{4}$
\thanks{$^{1}$Lena M. Downes is with the Department of Aeronautics and Astronautics,
        Massachusetts Institute of Technology, Cambridge, MA 02139, USA, and
        is a Draper Scholar with the Perception and Embedded ML Group,
        Draper, Cambridge, MA 02139, USA
        {\tt\small lmdownes@mit.edu}}%
\thanks{$^{2}$Dong-Ki Kim is with the Laboratory for Information and Decision Systems,
        Massachusetts Institute of Technology, Cambridge, MA 02139, USA
        {\tt\small dkkim93@mit.edu}}%
\thanks{$^{3}$Ted J. Steiner is with the Perception and Embedded ML Group,
        Draper, Cambridge, MA 02139, USA
        {\tt\small tsteiner@draper.com}}%
\thanks{$^{4}$Jonathan P. How is with the Faculty of Aeronautics and Astronautics,
        Massachusetts Institute of Technology, Cambridge, MA 02139, USA
        {\tt\small jhow@mit.edu}}%
        \thanks{Research funded by Draper}%
}

\begin{document}

\maketitle
\thispagestyle{empty}
\pagestyle{empty}

\begin{abstract}
Cross-view image geolocalization provides an estimate of an agent's global position by matching a local ground image to an overhead satellite image without the need for GPS. It is challenging to reliably match a ground image to the correct satellite image since the images have significant viewpoint differences. Existing works have demonstrated localization in constrained scenarios over small areas but have not demonstrated wider-scale localization. Our approach, called Wide-Area Geolocalization (\sysname{}), combines a neural network with a particle filter to achieve global position estimates for agents moving in GPS-denied environments, scaling efficiently to city-scale regions. \sysname{} introduces a trinomial loss function for a Siamese network to robustly match non-centered image pairs and thus enables the generation of a smaller satellite image database by coarsely discretizing the search area. A modified particle filter weighting scheme is also presented to improve localization accuracy and convergence. Taken together, \sysname{}'s network training and particle filter weighting approach achieves city-scale position estimation accuracies on the order of 20 meters, a 98\% reduction compared to a baseline training and weighting approach. Applied to a smaller-scale testing area, \sysname{} reduces the final position estimation error by 64\% compared to a state-of-the-art baseline from the literature.  \sysname{}’s search space discretization additionally significantly reduces storage and processing requirements.  A video highlight demonstrating particle filter convergence results for \sysname{} compared to the baseline for the Chicago test area is available at \url{https://youtu.be/06MOR0ozQeI}.
\end{abstract}

\section{INTRODUCTION}
\begin{figure}[t!]
  \centering
  \includegraphics[width=\linewidth]{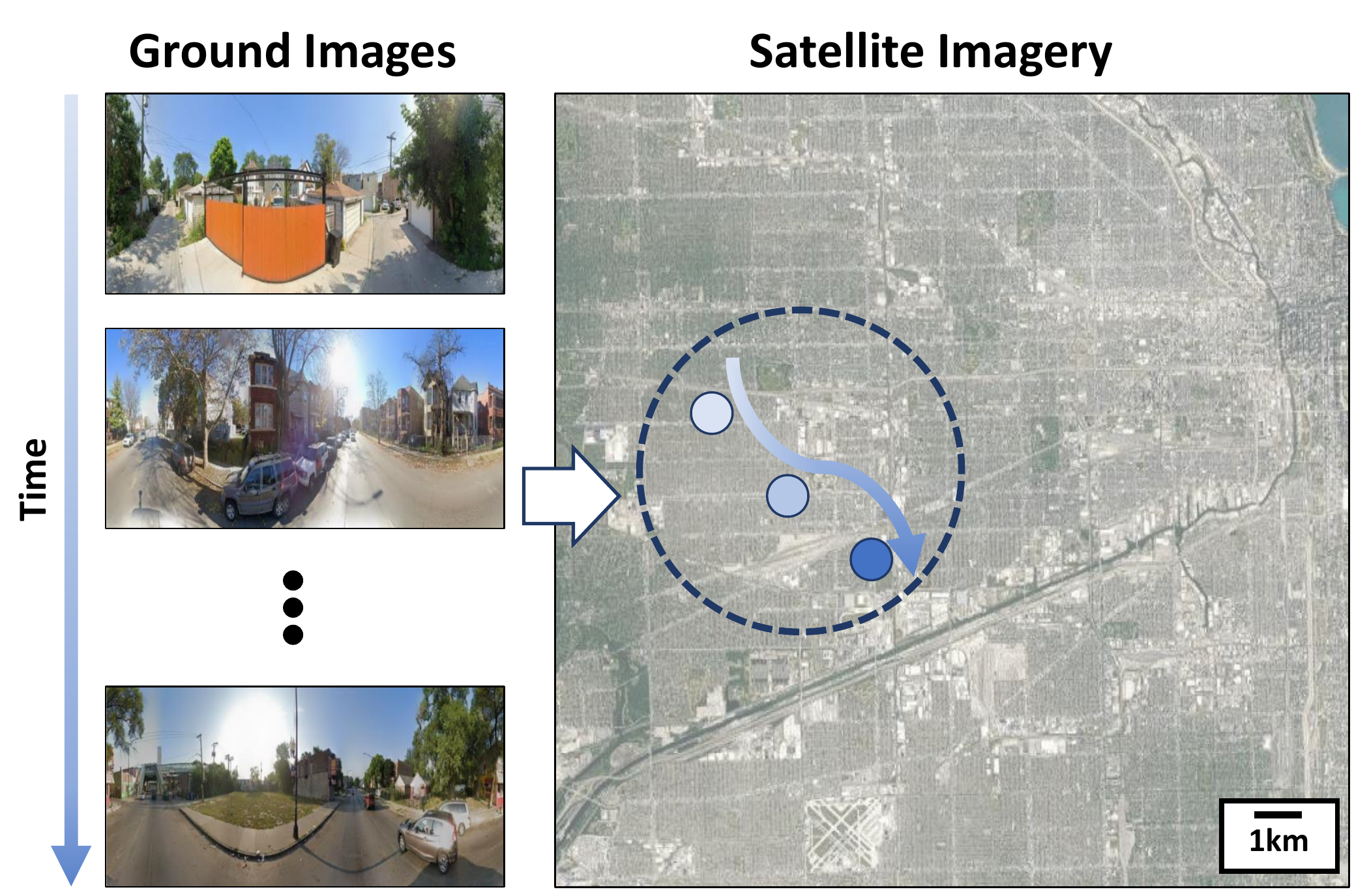}
  \caption{A ground-to-aerial geolocalization system takes in a series of ground-view camera images and satellite imagery of the search area and determines the agent location. \sysname{} enables an accurate localization on an extensive testing area that was not possible with prior works.}
  \label{fig:intro}\vspace*{-.2in}
\end{figure}

There are many scenarios in which GPS is unavailable or untrustworthy, such as urban canyons where building height causes GPS signals to be blocked, dropout, and when GPS is jammed or spoofed. A supplement to GPS that can provide accurate global localization is desirable so that poor GPS coverage can be enhanced or replaced. A promising solution to this problem is ground-to-aerial geolocalization (GTAG), shown in Fig.~\ref{fig:intro}, which matches ground-view camera images to aerial satellite images of the same location \cite{Cai, Tian, Hu, Kim, Shi}. GTAG effectively addresses the GPS issue because these image matches alone  generate an accurate location estimate over time. Satellite imagery is widely available, even in areas without reliable GPS coverage, and it is the only outside data needed for GTAG.

GTAG is a challenging problem due to the large viewpoint difference between the ground and satellite image views. Although significant advances have been made in recent years thanks to the introduction of deep learning to the problem \cite{Kim, Hu, Cai, Liu2019, Zhu2021, ZhuVIGOR}, image retrieval alone only accurately matches to the correct satellite image approximately 50\% of the time. Hence, recent works \cite{Xia, ZhuVIGOR, Hu, Kim} leverage a temporal localization aspect to improve localization performance beyond that which is achieved solely with image retrieval. However, these works require many constraints: access to noisy GPS data along the whole test path \cite{Xia, ZhuVIGOR}, perfect knowledge of initial location \cite{Hu}, search area constrained to less than 2.5 km$^2$ \cite{Kim, Hu}. These works cannot scale to accurate localization across an entire city with limited initialization because of the increased difficulty in matching imagery against a larger satellite database, demonstrated by ~\cite{Viswanathan}. The constraints of these works all decrease the effective size of the satellite database to enable accurate localization. City-wide GTAG requires a larger satellite image database to match against, which would cause existing works to experience performance drops. 

\textbf{Our contribution.} Our approach, Wide-Area Geolocalization (\sysname{}), makes several important changes to existing GTAG methods that enables city-wide localization on a scale that previous systems cannot demonstrate. First, \sysname{} maintains a coarser satellite image database instead of a dense database as required by previous methods. This is achieved by \sysname{}'s additional training with a trinomial loss to robustly match non-centered satellite images. As a result, \sysname{} effectively covers a large testing area, but also has faster computation and reduced storage. Second, \sysname{} improves a particle filter algorithm by using a Gaussian measurement model that incorporates the maximum satellite image similarity from the entire database. We observe that this change more accurately models the noise in the similarity measurements. These changes in total enable \sysname{} to produce more accurate localization across a larger area with fewer constraints than previous systems. In summary, we demonstrate the following benefits of \sysname{} in this paper:
\begin{enumerate}[leftmargin=*,topsep=-1em]
    \item reduces average and final estimation error,
    \item reduces convergence time of particle filter, and
    \item reduces storage and computational burden.
\end{enumerate}

\section{RELATED WORKS}

\textbf{Ground-to-aerial image matching.} The area of ground-to-aerial image matching pushes the limits of previous image matching solutions because it involves very different viewpoints. Initial works attempted to address this challenge using traditional computer vision techniques \cite{Viswanathan, viswanathan2016, jacobs, bansal}, but recent deep learning-based approaches have vastly improved the matching performance \cite{Workman}. Siamese networks \cite{bromley} (i.e., two jointly trained networks that produce an embedding scheme; also known as metric learning) are frequently used for this ground-to-satellite image matching\cite{Kim, Hu, Cai, Vo, Rodrigues, Shi, Cao, Zhu2021, Liu2019, lin, lin2015}. Typical approaches are based on the goal of one-to-one image retrieval: given a ground image and a database of satellite images, determine which satellite image is the best match. Performance is measured as the retrieval on top-k, which is the fraction of ground images that successfully rank their matching satellite image in the top-k of the entire satellite image database. Recall at top-1 lies at around 50\% \cite{Zhu2021}, which means that image retrieval alone can only match a ground image to the correct satellite image 50\% of the time. With 50\% recall at top-1, we cannot trust that a single image retrieval gives us the correct satellite image match. One effective way to determine an accurate location estimate is to combine the information from many different image matches over time. Unlike those image retrieval focused systems, \sysname{} is designed to maximize the localization information gained from combining image matches over time.

\textbf{Particle filter implementations.} Several previous works have used a particle filter to address the need for temporal localization\cite{Kim,Hu,Xia, Viswanathan, viswanathan2016}. Particle filters use random discrete particles to estimate a probabalistic distribution of a state hypothesis given control inputs and sensor measurements. In GTAG, particle filters estimate the agent location given odometry information and similarity measurements from the Siamese network. Using a particle filter improves the localization performance over using image retrieval alone, but previous works that have used this approach require dense sampling of the localization area into satellite images, which generates a large database -- (e.g., a satellite image for every 5~m \cite{Hu} or one for each particle \cite{Kim}).  Those approaches must sample satellite images densely because their neural networks are designed assuming that all ground images will be centered within the matching satellite image. The large number of satellite images generated from dense sampling makes city-wide localization infeasible due to the computational and storage burden, since an embedding must be stored and a similarity must be calculated for each database image. In contrast, \sysname{} requires a much smaller satellite image database by sampling satellite imagery more coarsely and hence facilitates city-wide localization with little computational and storage burden.

\section{METHODS}
\subsection{Overview of Approach}
\begin{figure*}[t!]
\centering
  \includegraphics[width=0.8\linewidth]{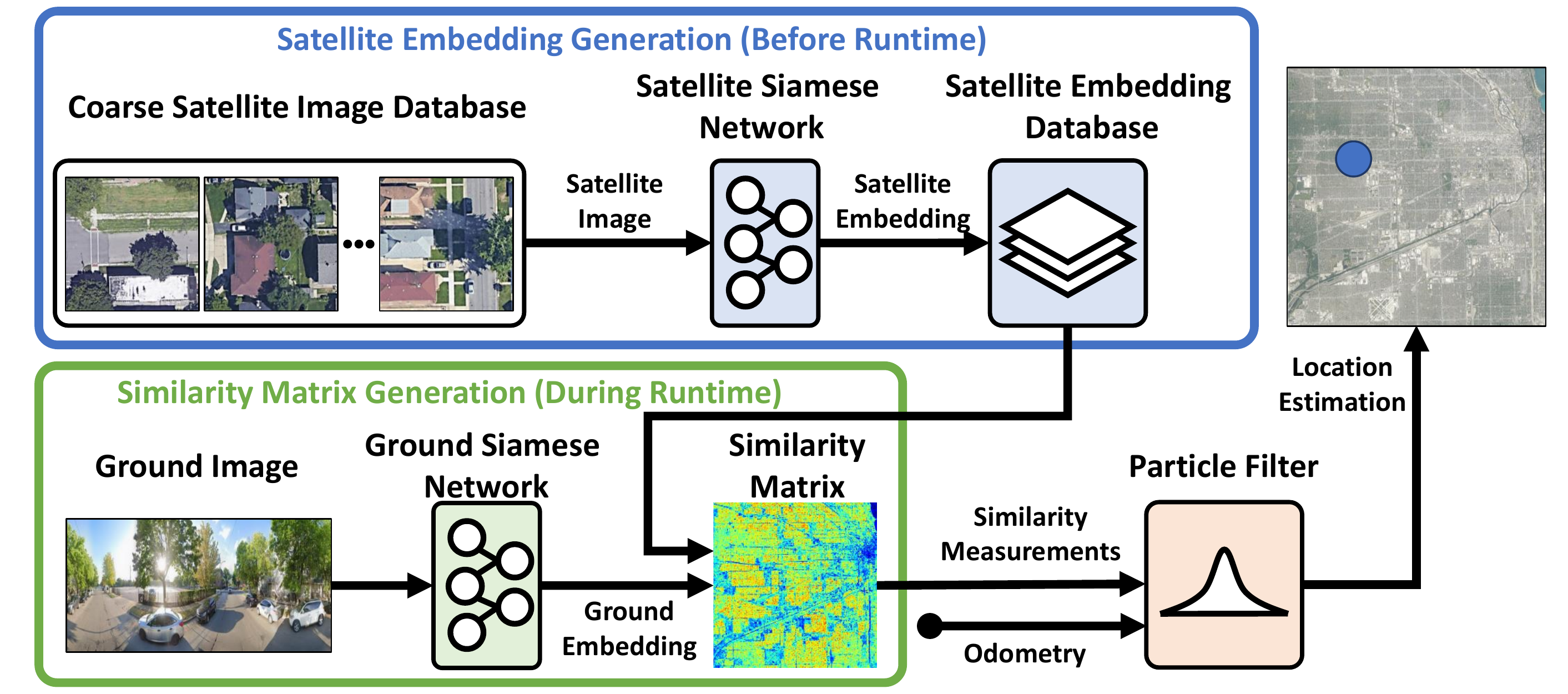}
  \caption{Diagram of \sysname{}. Satellite embeddings are generated before runtime with the coarsely sampled satellite image database and the Siamese network that was trained with trinomial loss. During runtime, the ground embedding is generated at each time step and combined with the satellite embeddings to create a similarity matrix. Odometry and measurements from the similarity matrix are input to the particle filter with a Gaussian measurement model to generate a location estimate.}
  \label{fig:system}
\vspace*{-0.2in}
\end{figure*}

\sysname{} mainly consists of a strategy for creating a satellite image database, a Siamese network, and a particle filter (see Fig.~\ref{fig:system}). \sysname{} first constructs a database of satellite images by coarsely sampling the entire search area. \sysname{} then uses the Siamese network to preprocess all satellite images in the database before runtime to accelerate measurement updates. During runtime, the ground image is embedded by the Siamese network and the similarity between the ground embedding and each satellite embedding is calculated, producing a similarity matrix. Then, the particle filter receives measurements from the similarity matrix and odometry as input to produce a location estimate at each time step. Now, we describe each sub-system in detail.

\subsection{Satellite Image Database}
\begin{figure}[t!]
\centering
  \includegraphics[width=0.9\linewidth]{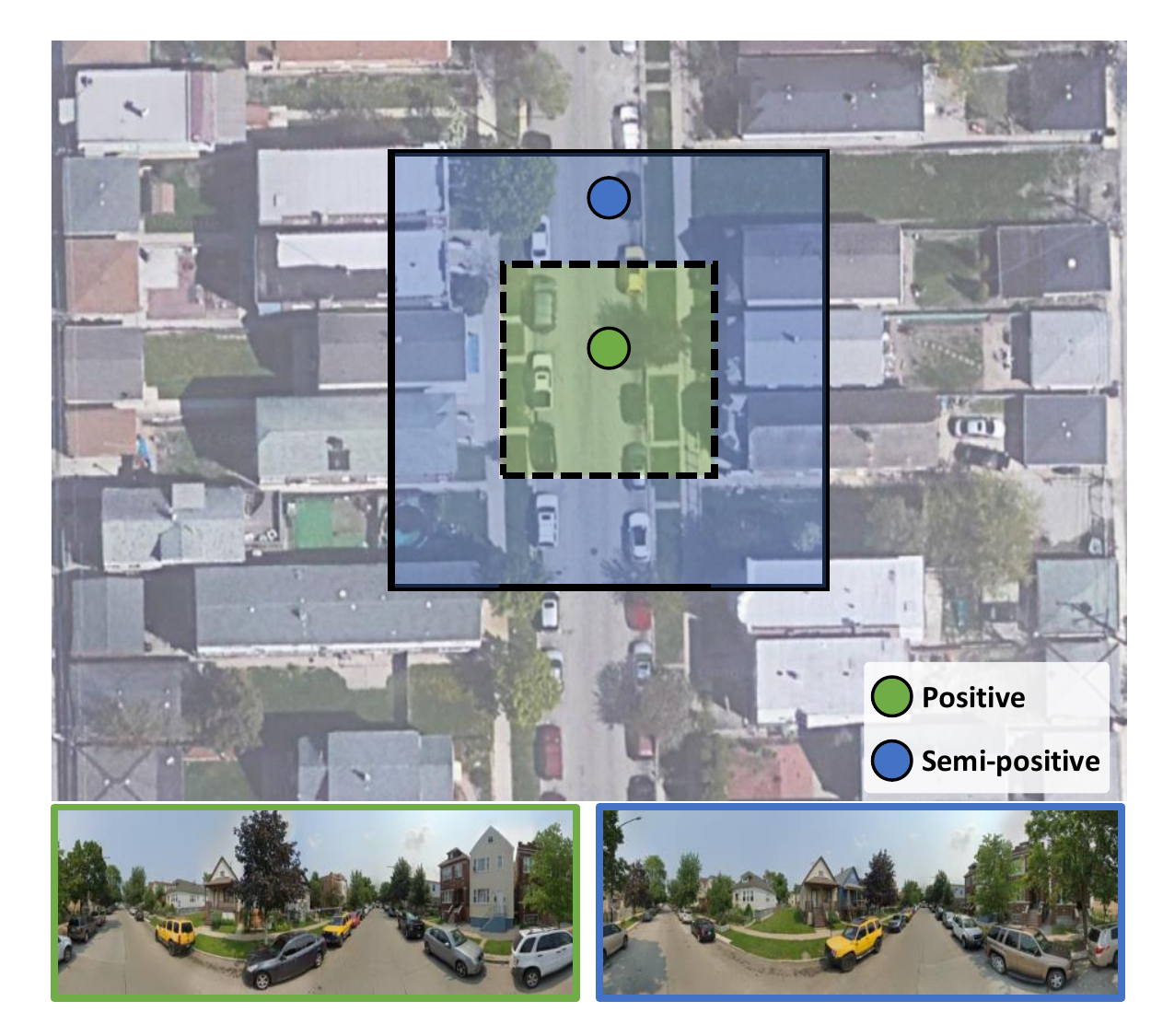}
  \caption{Positive pairs vs. semi-positive pairs. For the satellite image outlined in solid black, a ground image taken in the green shaded center region (ground image shown outlined in green) makes a positive pair and a ground image taken in the blue shaded region (ground image shown outlined in blue) makes a semi-positive pair.}
  \label{fig:pos_semi}\vspace*{-0.2in}
\end{figure}

Previous works require dense sampling of the localization area for satellite images \cite{Hu, Kim} because of how their neural networks were trained. In particular, \cite{Hu, Kim} require specific constraints on where the ground images are taken, such as requiring ground images to be centered within their matching satellite images. Figure~\ref{fig:pos_semi} shows the difference between positive (centered) image pairs and semi-positive (non-centered) pairs. In previous works, each ground image could only be matched to the positive part of a satellite image-- the green middle portion of Fig.~\ref{fig:pos_semi}. Hence, the satellite database must contain enough images to cover the search area in green, significantly increasing the number of satellite images needed for localization within an area. However, requiring such dense sampling of satellite images increases the amount of computation that must be done to match each ground image to a satellite image, resulting in a system that cannot function close to real-time on large localization areas. 

Instead, we took an alternative approach that coarsely samples the localization area and enables \sysname{} to also use the blue portion of Fig.~\ref{fig:pos_semi} for localization. Specifically, motivated by \cite{ZhuVIGOR}, we train our Siamese network to match both centered (positive) image pairs and non-centered (semi-positive) image pairs, so that all of the area within a satellite image can be matched against. Thus \sysname{} can fully utilize each satellite image and does not need to cover the search area with the positive portion of satellite images. As a result of this process we can generate a database with much fewer images.

\textbf{Trinomial loss}. \sysname{} has additional network training to improve its performance on non-centered semi-positive pairs and therefore enable coarser sampling of satellite images. We take the binomial loss of \cite{Zhu2021} one step further by adding a term for semi-positive pairs in Eq.~\ref{eq:Ls}:
\begin{equation}\label{eq:Ls}
\mathcal{L}_{\text{semi}} = \frac{\log(1+e^{-\alpha_{\text{semi}}(S_{\text{semi}}-m_{\text{semi}})})}{N_{\text{semi}}\alpha_{\text{semi}}} \\
\end{equation}
\begin{equation}
\label{eq:Ltri}
\mathcal{L}_{\text{trinomial}} = \mathcal{L}_{\text{binomial}} + \mathcal{L}_{\text{semi}}.
\end{equation}
Here $\alpha_{\text{semi}}$ is the weight that adjusts the magnitudes of the gradient for semi-positive pairs, $S_{\text{semi}}$ is the cosine similarity of the pair of semi-positive embeddings, $m_{\text{semi}}$ is the average value of $S_{\text{semi}}$, and $N_{\text{semi}}$ is the number of semi-positive pairs. $ \mathcal{L}_{\text{binomial}}$ is the sum of a positive and negative loss term, fully defined in \cite{Zhu2021}. Works previous to \cite{ZhuVIGOR} treated semi-positive pairs as the same as negative pairs, since they assumed that each ground image would always have a perfectly centered satellite image that should be treated as its only match. This was inherent in the performance metrics used-- recall at top-k measures how similar the embedding of the one true centered satellite image is to the query ground image. Ref.~\cite{ZhuVIGOR} incorporated a new loss function to improve performance on semi-positive pairs, but our testing revealed its performance drops from 41.1\% recall on top-1 for positive image pairs to 3.7\% recall on top-1 for semi-positive image pairs. Our additional trinomial loss training improves performance on semi-positive pairs and enables coarse satellite image sampling in which a ground image will frequently not have a centered satellite match.  %

\subsection{Particle Filter Measurement Model}
\begin{figure}[t!]
\centering
  \includegraphics[width=0.7\columnwidth]{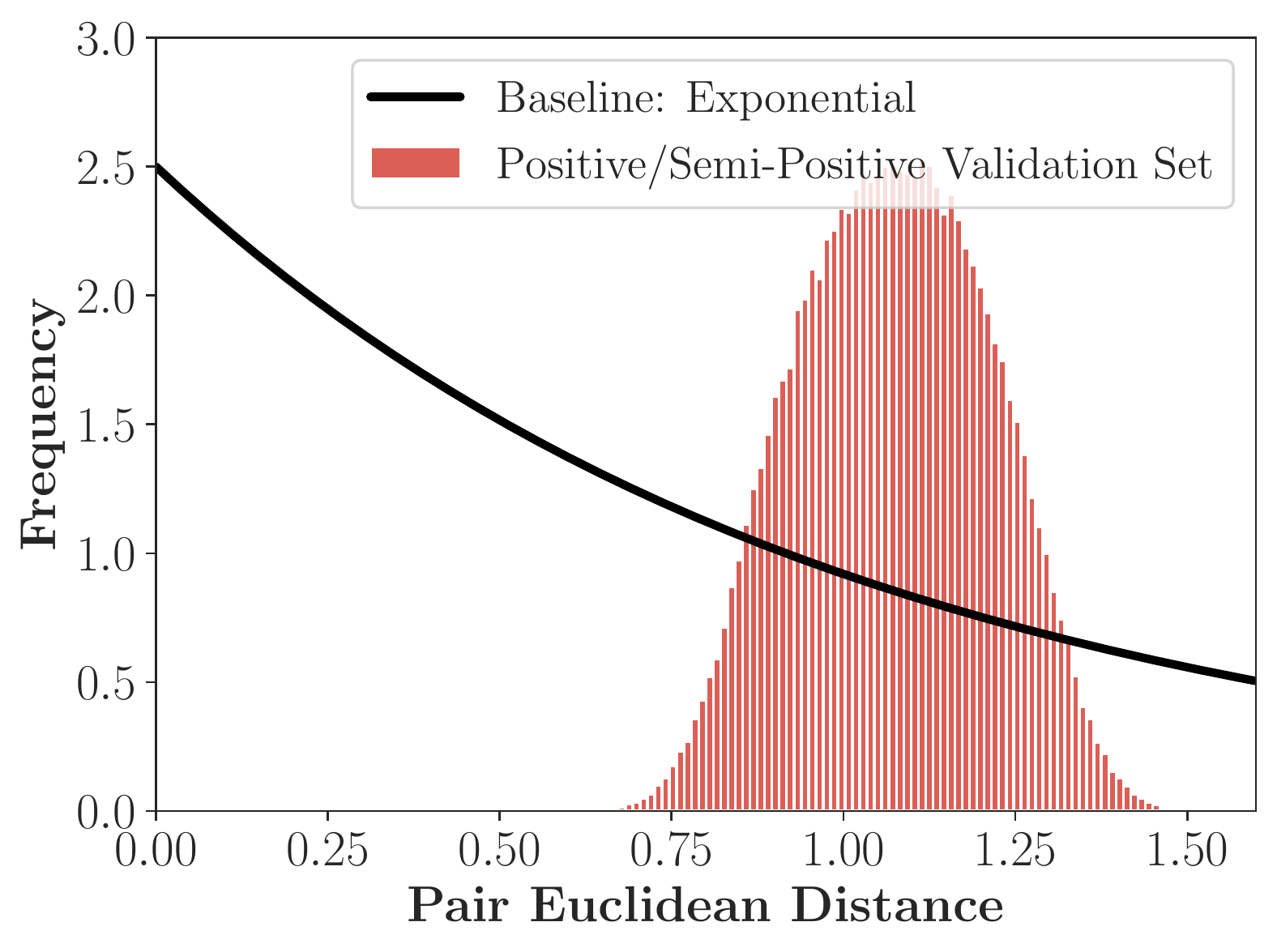}
  \vspace*{-0.15in}
  \caption{Baseline particle filter measurement model compared to empirical distribution of positive/semi-positive pair Euclidean distance measurements in the validation set from \cite{ZhuVIGOR}'s dataset. An exponential measurement model does not fit the data.}
  \label{fig:base_meas_model}
\end{figure}

\sysname{} also modifies the particle filter measurement model compared to prior work\cite{Hu,Kim}, which uses an exponential measurement model:
\begin{equation}
\label{eq:exp}
P(z_t|x_t) = \beta e^{-\beta d_{t}^{k}},
\end{equation} 
where $z_t$ is the sensor measurement for timestep $t$, $x_t$ is the agent pose at $t$, $d_{t}^{k}$ is the Euclidean distance between the embedding pair for satellite image $k$ and the ground image from $t$, and $\beta$ is a tuning parameter. The exponential measurement model assumes that as image pair Euclidean distance decreases, the probability of a match increases exponentially. However, our key observation is that an exponential measurement model does not generally match the empirical data distribution for Euclidean distance between positive and semi-positive pairs, as demonstrated in Fig.~\ref{fig:base_meas_model}. The data follows a Gaussian distribution, and the exponential measurement model does not properly account for noise in the neural network output. This mismatch causes wrong estimates of the likelihood of measurements, which affects particle filter convergence. 

\begin{figure}[t!]
\centering
  \includegraphics[width=0.7\columnwidth]{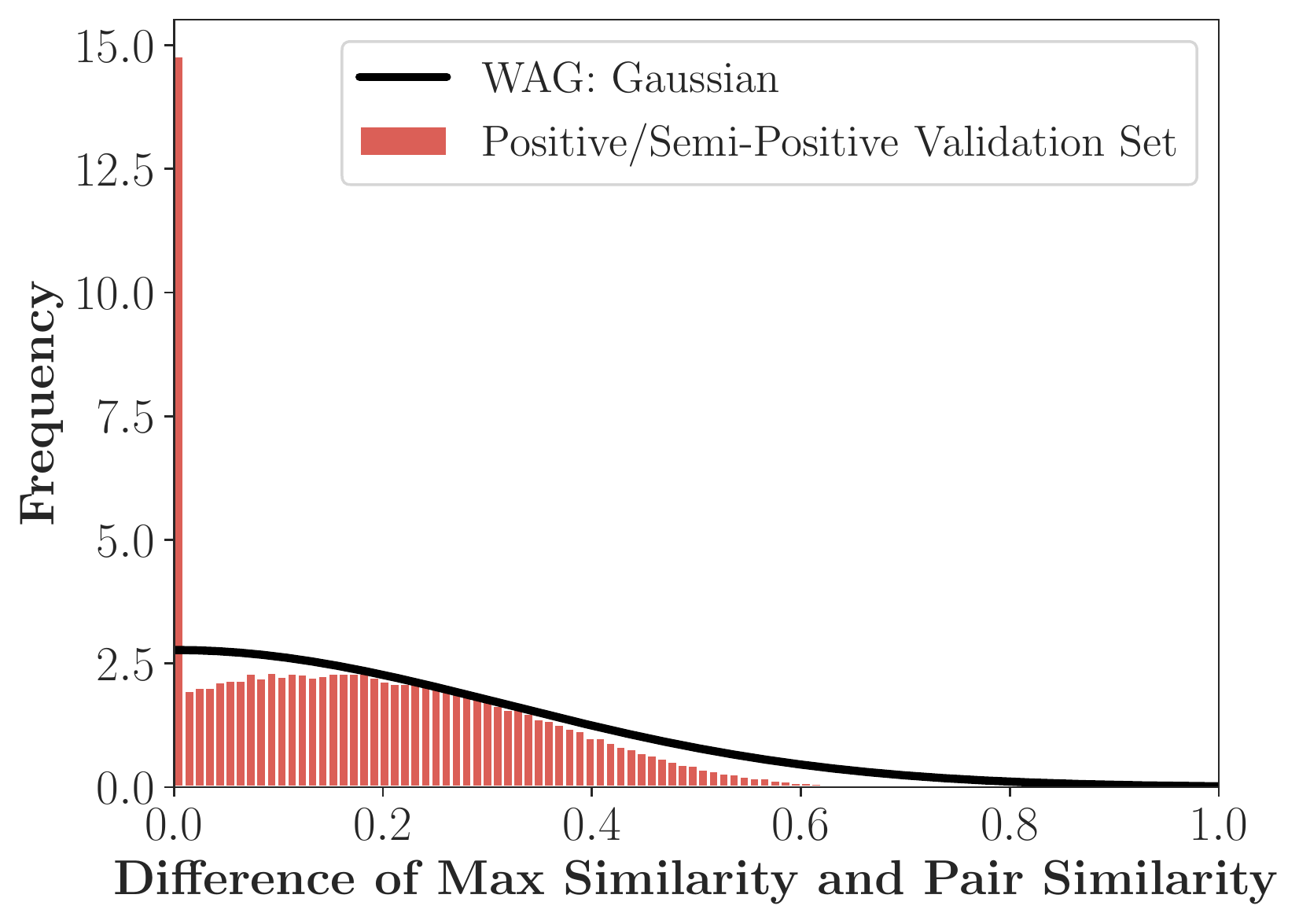}
  \vspace*{-0.15    in}
  \caption{\sysname{}'s particle filter measurement model of Eq.~\ref{eq:gauss} compared to empirical distribution in the validation set from \cite{ZhuVIGOR}'s dataset. \sysname{}'s measurement model better fits the data. Empirical data spike at 0 is due to some pairs being the database pair with maximum similarity.}
  \label{fig:our_meas_model}\vspace*{-0.2in}
\end{figure}

Given that the measurement for each ground-satellite pair is the difference between the maximum database similarity and the query pair similarity at that time step, 
($
z_t = \max(s_{t})-s_{t}^{k}$),
\sysname{} instead uses the Gaussian measurement model:
\begin{equation}
\label{eq:gauss}
P(z_t|x_t) = \frac{1}{\sigma\sqrt{2\pi}} 
  \exp\left( -\frac{1}{2}\left(\frac{\max(s_{t})-s_{t}^{k}}{\sigma}\right)^{\!2}\,\right),
\end{equation} 
which better captures the output distribution of the neural network. The covariance $\sigma$ was set to 0.1 to approximately fit the output distribution observed in the validation data for positive and semi-positive pairs shown in Fig.~\ref{fig:our_meas_model}. \sysname{} also uses cosine similarity instead of Euclidean distance. In addition, the use of $z_t$ as the measurement causes the distribution to be zero mean, which removes the need to find the distribution mean, another tuning parameter.

\section{RESULTS}
\subsection{Experimental Setup}
We perform localization experiments in two settings: a large scale localization with very noisy location initialization across the entire city of Chicago, and a small-scale localization with perfect location initialization across a neighborhood in Singapore. These were experiments with simulated data: panoramic ground images and overhead satellite images from Google Maps Static API and odometry measurements from the ground-truth displacement between images with added noise proportional to displacement. In Chicago, the satellite imagery was sampled approximately every 60 meters into a 256 $\times$ 256 grid of non-overlapping satellite image tiles. In Singapore, the satellite imagery was sampled approximately every 90 meters into a 16 $\times$ 16 grid. These grid sizes were chosen to maintain the same image size and resolution that the network was trained on; satellite images of size 640 $\times$ 640 pixels with a resolution of approximately 0.1 m/pixel.

Both experiments use the neural network in~\cite{ZhuVIGOR} trained with trinomial loss from epoch 30 for an additional 15 epochs with the values in Table~\ref{tab:loss_params}. The filter in both settings has 100,000 particles and uses the Gaussian measurement model in Eq.~\ref{eq:gauss}. In the Chicago experiment, the particle filter was initialized with a Gaussian distribution centered 1.3 km from the true location (standard deviation of 2.97 km). In the Singapore experiment, the particle filter was initialized at the ground-truth location exactly. The Chicago experiment added 2\% noise to the ground-truth odometry and the Singapore experiment added 5\% odometry noise.

\begin{table}[t]
\vskip+0.1in
\caption{Trinomial Loss Parameters}
\vspace*{-0.2in}
\label{tab:loss_params}
\begin{center}
\begin{tabular}{|l|c|c|}\hline
\bf Parameter & \bf Symbol & \bf Value \\\hline
Positive pair weight &$\alpha_{p}$  & 5 \\
Semi-positive pair weight &$\alpha_{s}$    & 6 \\
Negative pair weight &$\alpha_{n}$    & 20 \\
Positive pair average similarity &$m_{p}$  & 0 \\
Semi-positive pair average similarity &$m_{s}$    & 0.3 \\
Negative pair average similarity &$m_{n}$    & 0.7 \\\hline
\end {tabular}
\end{center}
\vspace*{-0.3in}
\end{table}

\subsection{Computation and Storage}
\begin{figure}[t!]
\centering
  \includegraphics[width=0.7\columnwidth]{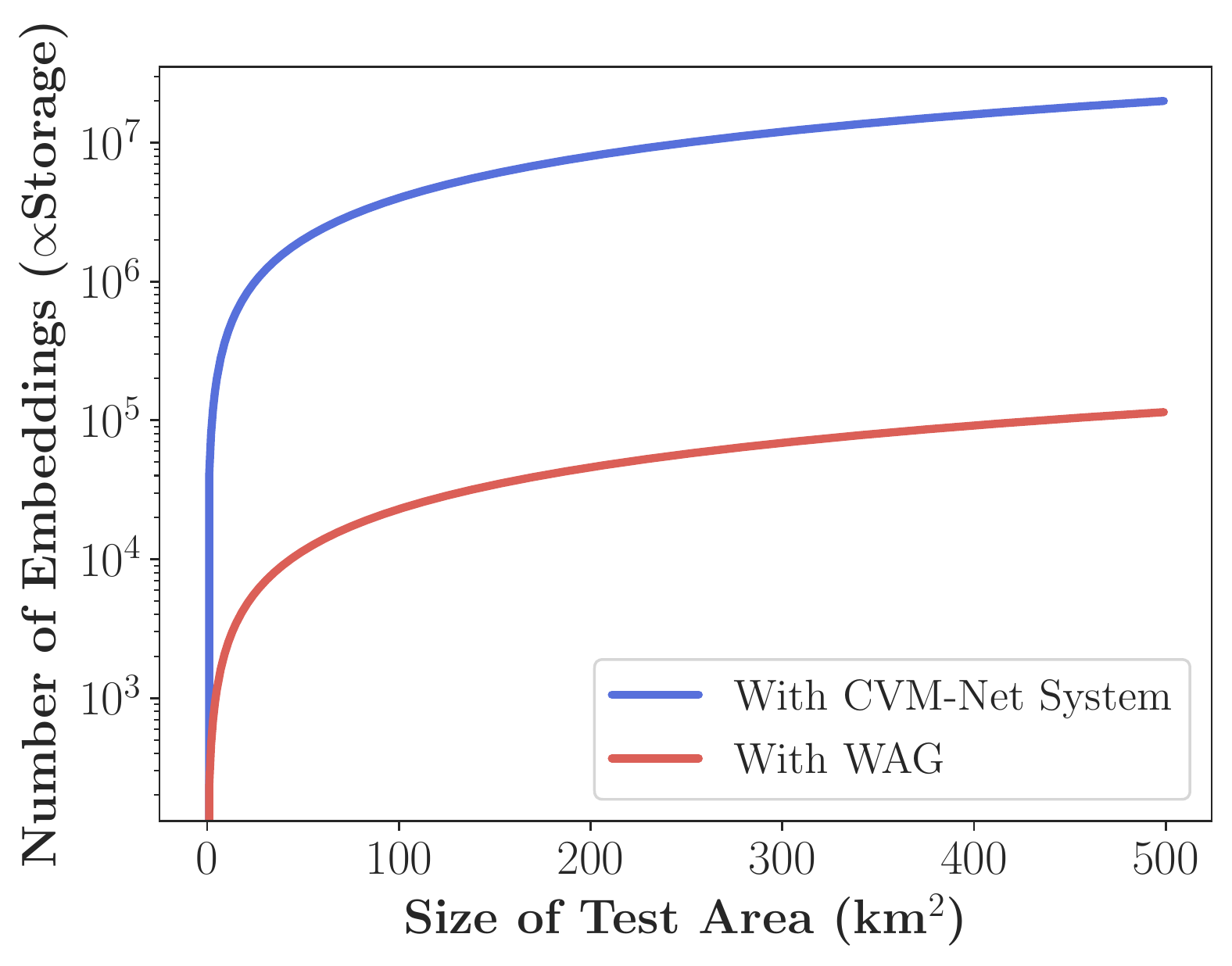}
  \vspace*{-0.15in}
  \caption{Storage requirements of \sysname{} compared to those of CVM-Net \cite{Hu}. \sysname{}'s storage requirements grow very slowly in comparison. For \sysname{} we assume a satellite image sampling density of 1 image per 66 m as used in the Chicago test, and for CVM-Net we assume 1 image per 5 m as specified in \cite{Hu}.}
  \label{fig:storage}
\centering
  \includegraphics[width=0.7\columnwidth]{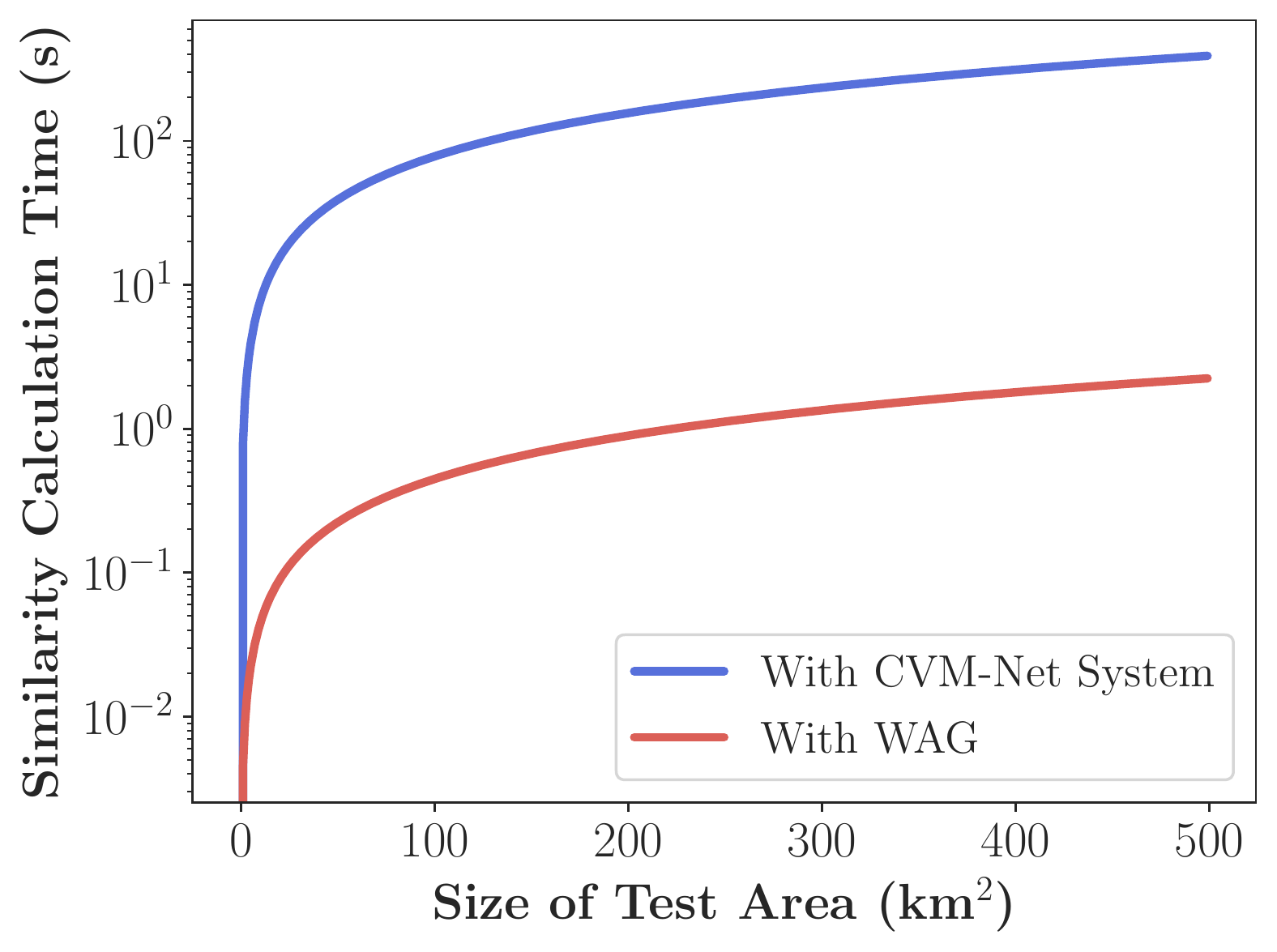}
  \vspace*{-0.15in}
  \caption{Computation requirements of \sysname{} compared to that of CVM-Net \cite{Hu} for calculation of the similarity matrix at each time step of the particle filter, with same sampling assumptions as in Fig.~\ref{fig:storage}. \sysname{}'s computation requirements are feasible for real-time usage, CVM-Net is largely infeasible for real-time usage once a test area is larger than 10 km$^2$.}
  \label{fig:computation}
\end{figure}

Our ability to do coarser satellite image sampling provides greater flexibility for computation and storage and enables a wider variety of applications for robotics. Since previous systems required all ground images to be centered within their satellite image pairs, for a given test area they required many satellite images, hence many satellite image embeddings had to be computed and stored before runtime (see Fig.~\ref{fig:storage}). 

The number of satellite images also affects computation because at each time step the similarity must be calculated between the ground embedding and every satellite embedding. Figure~\ref{fig:computation} demonstrates the timing required for a single particle filter measurement update with $1.9552\times 10^{-5}$ s as the estimated time for each similarity calculation (the average for our system). \sysname{} has a similarity matrix calculation time of $\approx$1.28 seconds for the Chicago test area of nearly 300 km$^2$, while CVM-Net would take $\approx$3.8 minutes. \sysname{}'s timing in Chicago would make real-time localization feasible as that test path does a measurement update approximately every 250 m, which would likely take at least 10 seconds for any ground agent to travel in a city environment.
%


\subsection{Large-scale Test: Chicago} 
\begin{figure}[t!]
\centering
  \includegraphics[width=0.9\linewidth]{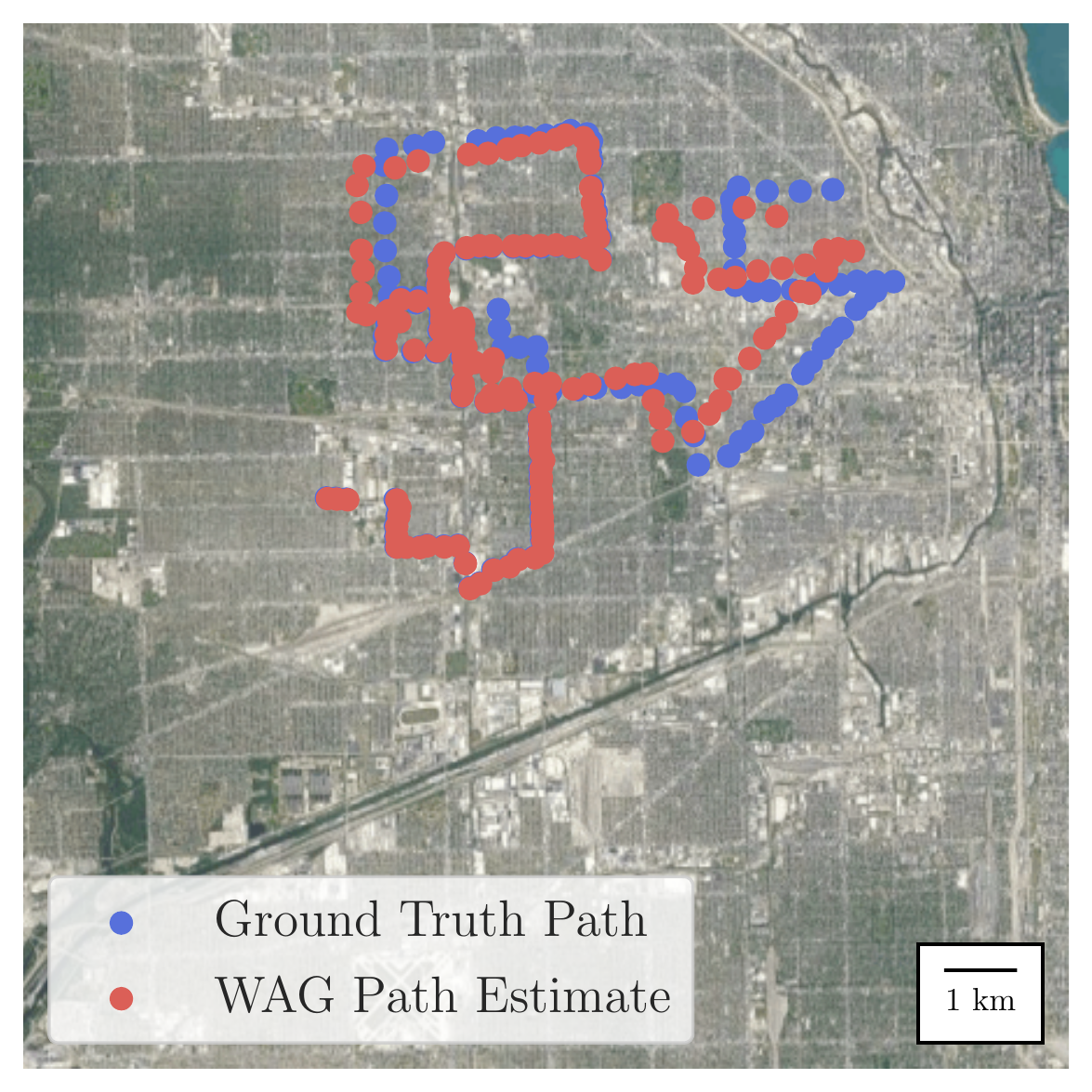}
  \vskip-0.15in
  \caption{The ground-truth path in Chicago and \sysname{}'s particle filter path estimate, which accurately converges upon the ground-truth over time.}
  \label{fig:path}
\end{figure}

\textbf{Experiment details.} To demonstrate \sysname{}'s ability to localize within a very large area, we complete a localization experiment on the scale of the entire city of Chicago. The true simulated path is shown in Fig.~\ref{fig:path}, overlaid with our estimated location at each time step. We ran this experiment with \sysname{} and a baseline that is similar to \sysname{}, but which is not trained with trinomial loss and which uses the measurement function in Eq.~\ref{eq:exp} in its particle filter (consistent with the approach in \cite{Kim,Hu}) instead of the proposed one in Eq.~\ref{eq:gauss}.

\begin{figure}[t!]
  \centering
  \subfigure[Initial distribution supplied to particle filter. True location is over 1 km from initial particle filter estimate. \label{fig:initial_gauss}]{\includegraphics[width=.49\columnwidth]{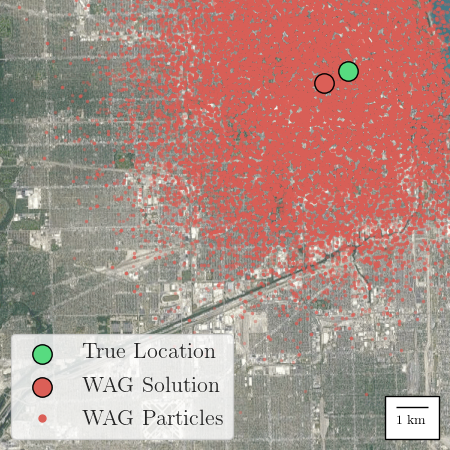}}
  \subfigure[Final particle filter solution and true location. True location is approximately 21 m from final particle filter estimate. \label{fig:zoom_nopart}]{\includegraphics[width=.49\columnwidth]{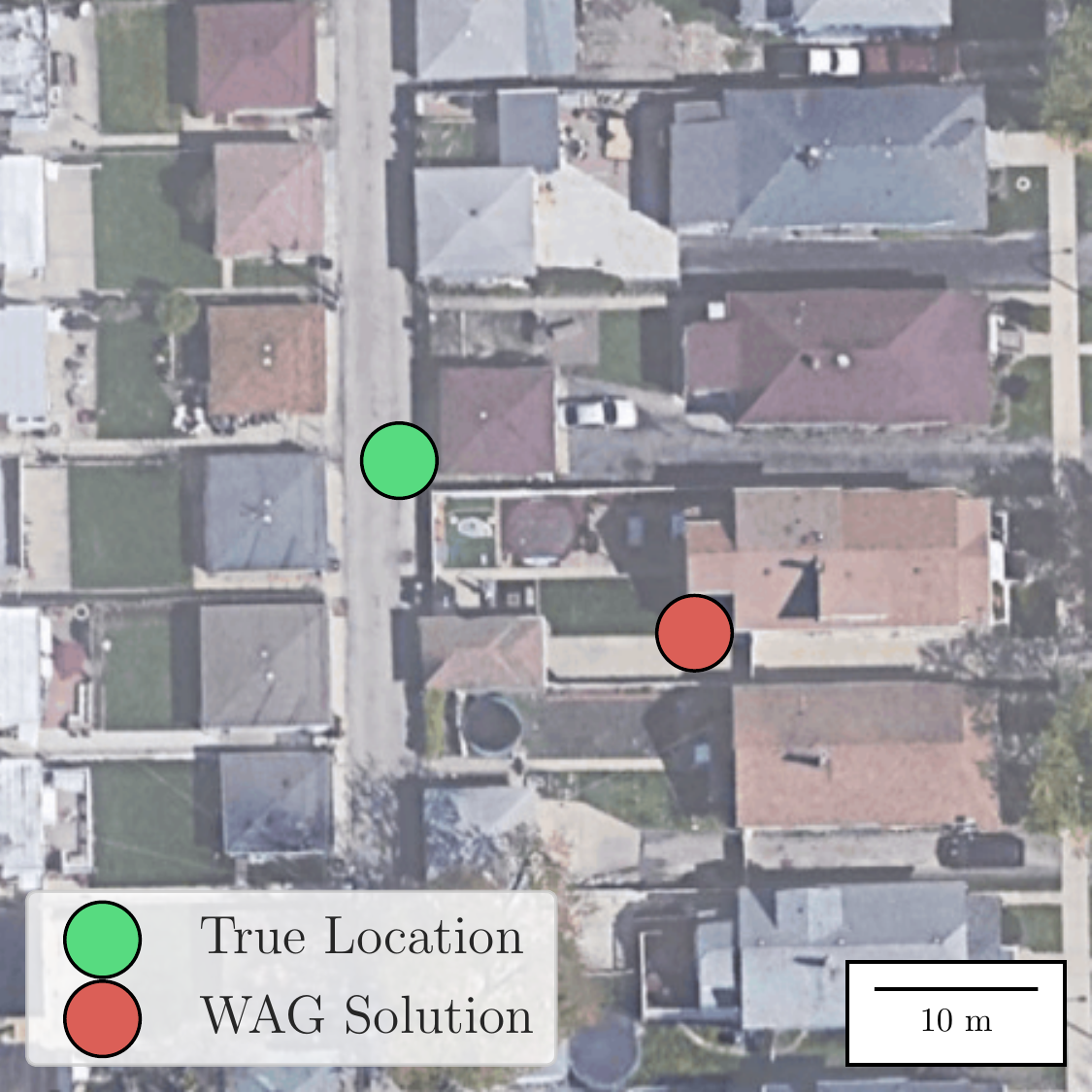}}
  \caption{\sysname{} is able to accurately localize an agent to within 21 meters of its true location across nearly 200 km$^2$ of Chicago after being initialized to a Gaussian distribution centered 1.3 km from the true location.}
  \vspace*{-0.2in}
  \label{fig:bam}
\end{figure}

\begin{figure}[t]
\centering
  \includegraphics[width=0.8\linewidth]{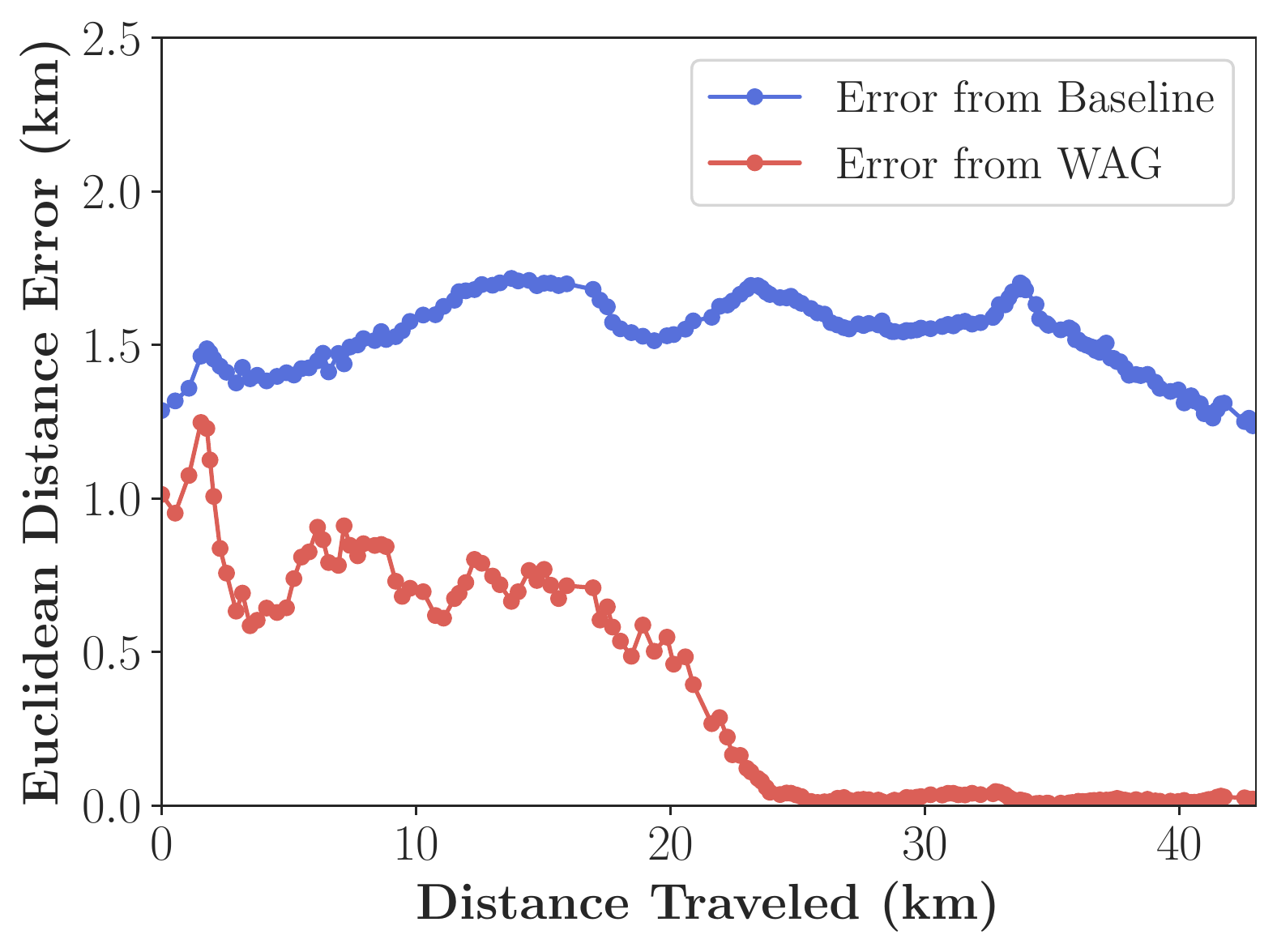}
  \vspace*{-0.2in}
\caption{Particle filter estimation error from \sysname{} compared to baseline system. Baseline is without trinomial loss training and with exponential measurement model. \sysname{} has lower final and average error.}
  \label{fig:err_conv}
\centering
  \includegraphics[width=0.8\linewidth]{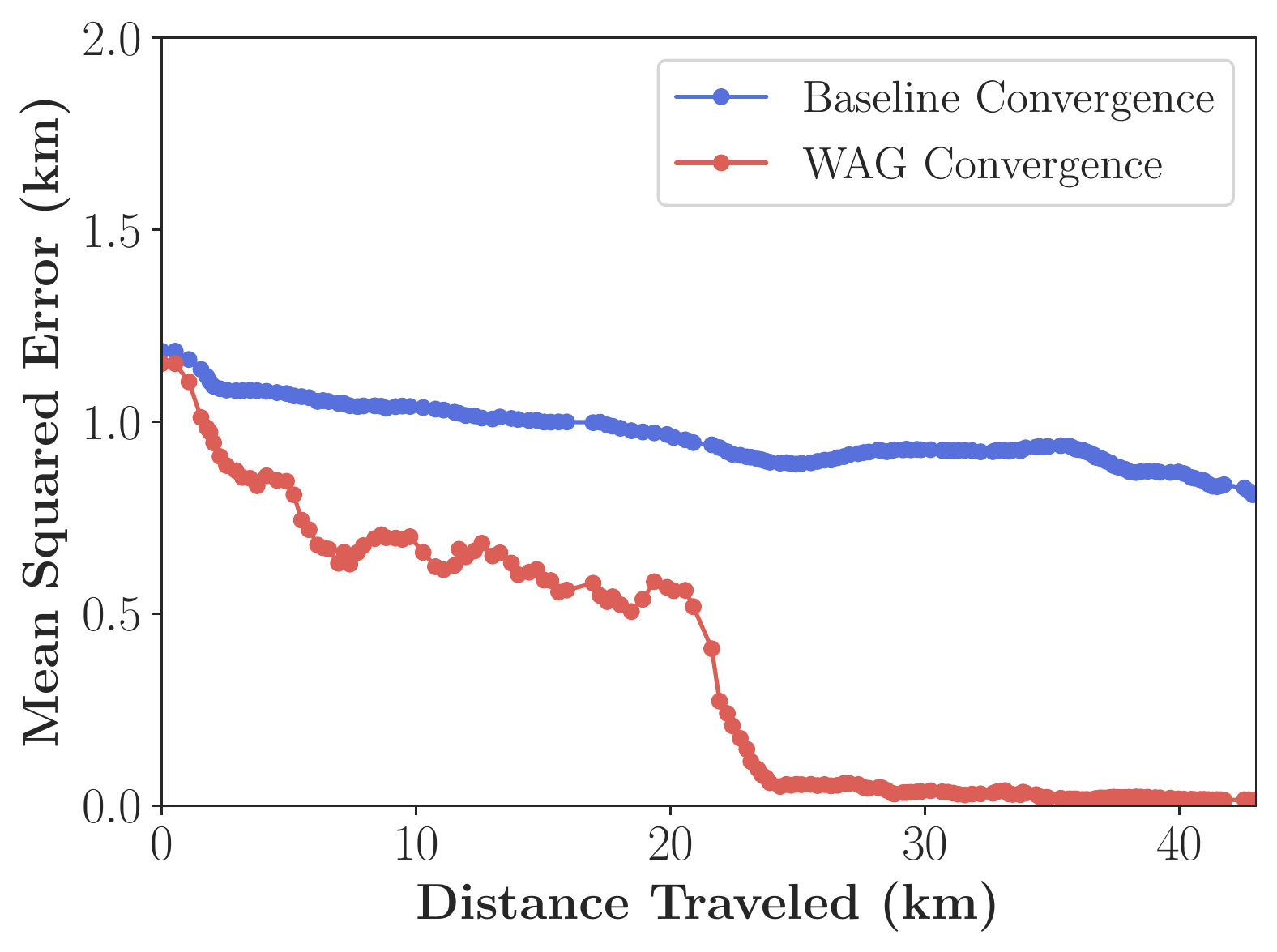}
  \vspace*{-0.2in}
\caption{Particle filter estimation convergence from \sysname{} compared to baseline. Baseline is without trinomial loss training and with exponential measurement model. \sysname{} converges to an accurate estimate but the baseline does not.}
  \label{fig:mse_conv}\vspace*{-0.1in}
\end{figure}

\textbf{Estimation error.} Even with an initialization as far from the agent as Fig.~\ref{fig:initial_gauss}, \sysname{} has a final estimation error of 21 m (visible in Fig.~\ref{fig:zoom_nopart}) that is so small that it is comparable to the scale of the satellite tiles used for localization. Thus the estimation accuracy of WAG has essentially reached the noise floor of the system. Fig.~\ref{fig:err_conv} compares the estimation error of \sysname{}'s particle filter and the baseline as the simulated agent moves. The error is the Euclidean distance between the actual location and the output of the particle filter, which is the weighted average of the locations of the particles.  Over the time period shown, \sysname{} has an average estimation error of 314 m, versus 1.5 km for the baseline. The baseline has a final estimation error of 1236 m. To the best of the authors' knowledge, \sysname{} is the only GTAG system in the literature that localizes this accurately over such a large area.

\begin{figure}[t]
\centering
  \includegraphics[height=0.815\linewidth]{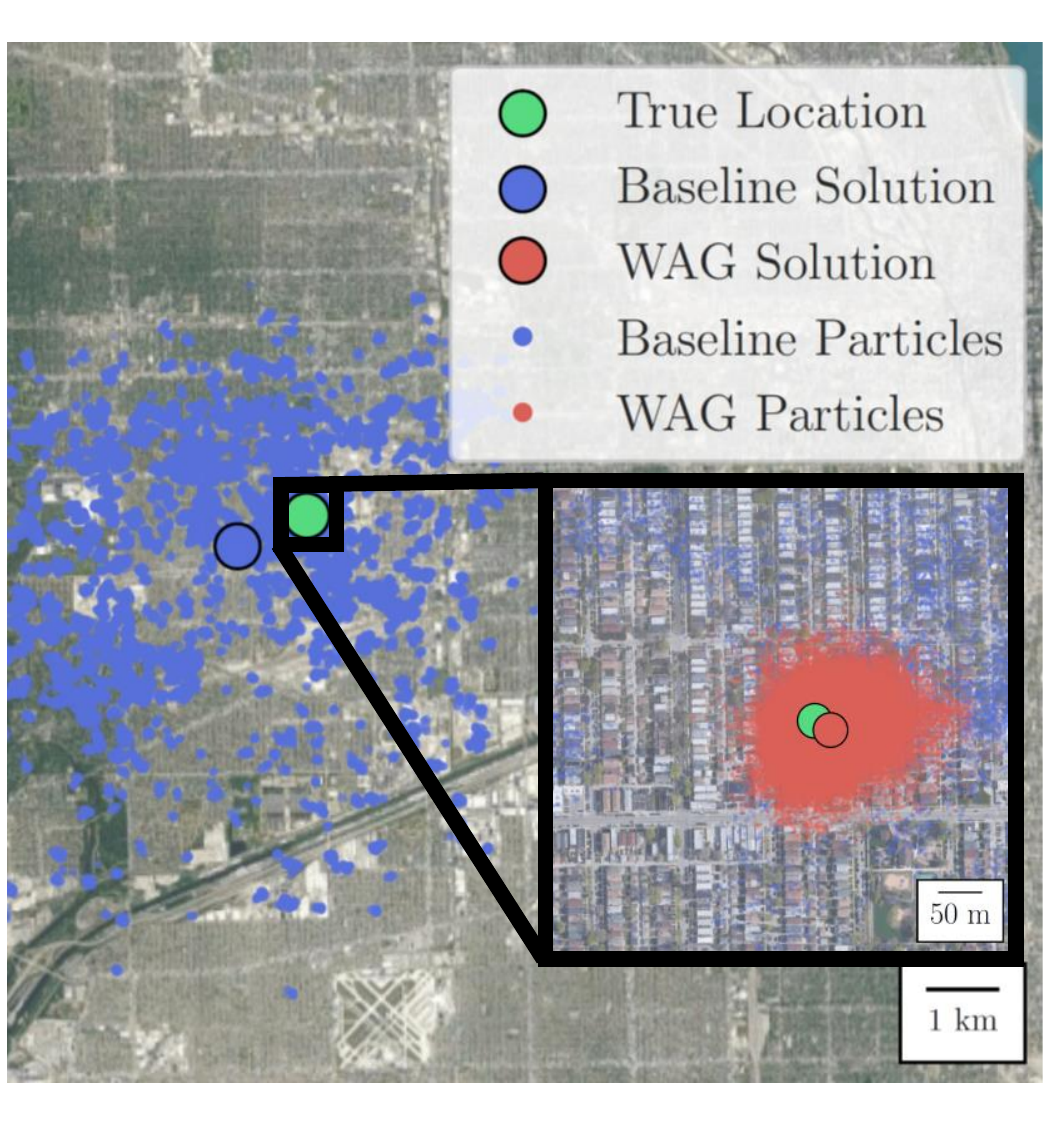}
  \vspace*{-0.1in}
\caption{Final particle filter dispersion with the baseline system and with \sysname{} on the Chicago test path (C-1). The baseline does not successfully converge to a location estimate, while \sysname{} converges to within 60 meters.}
  \label{fig:exp_pf}
%
\centering
  \includegraphics[width=0.8\linewidth]{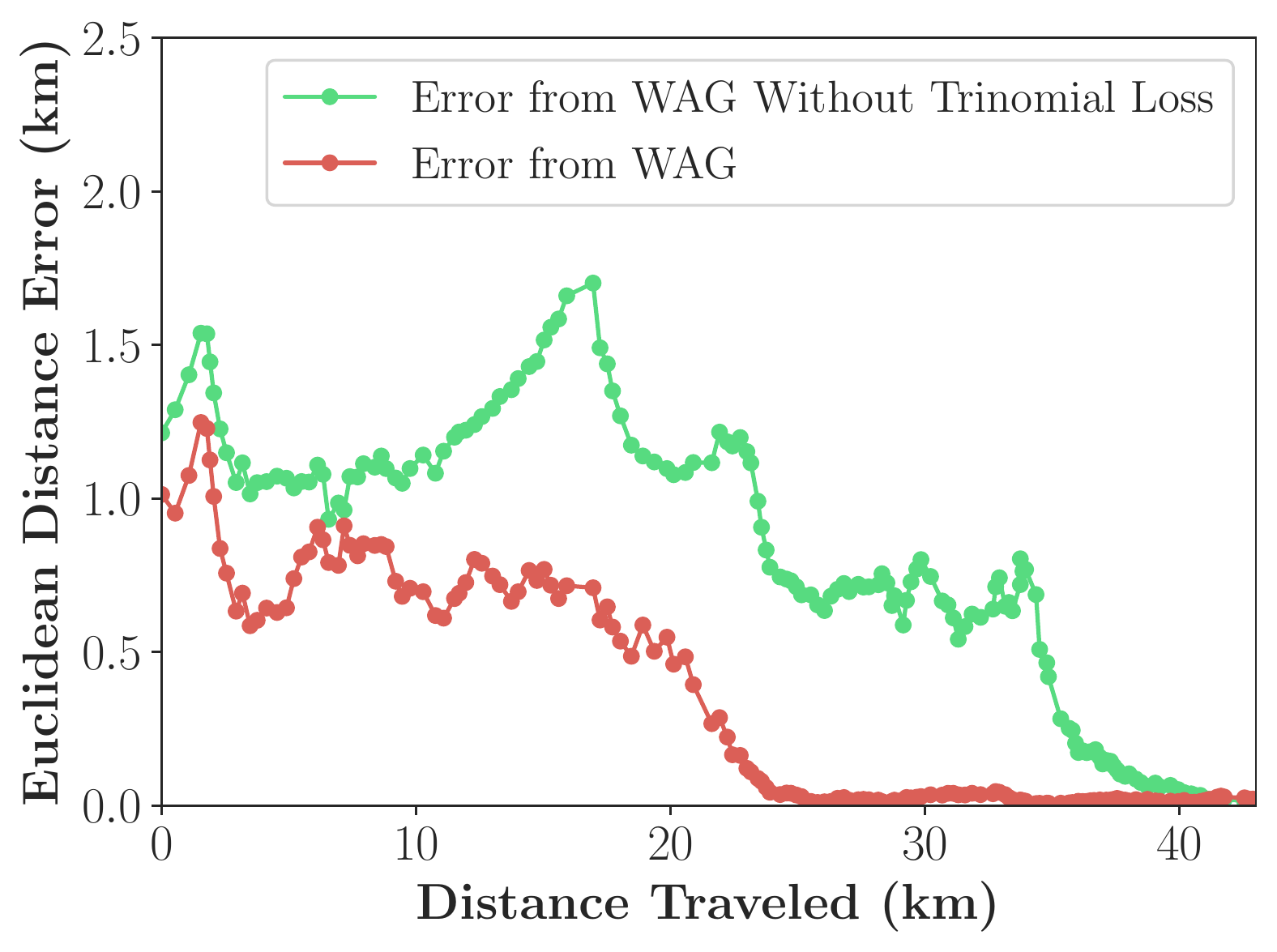}
\vspace*{-0.2in}
  \caption{The estimation error on the same test path using \cite{ZhuVIGOR}'s network without additional trinomial loss training}
  \label{fig:vig_conv}
  \vspace*{-.2in}
\end{figure}

\textbf{Convergence.} Figure~\ref{fig:mse_conv} compares the mean squared error of the particle locations at each time step, which can be viewed as a measure of convergence \cite{crisan}. \sysname{} converges to a mean squared error of less than 60 m (the satellite image size) after 74 filter updates, while the baseline does not reach that level of convergence in the time period tested. \sysname{} converges to the particle filter distribution shown by the red dots in the inset of Fig.~\ref{fig:exp_pf} while the baseline terminates with the particle distribution shown in Fig.~\ref{fig:exp_pf} (blue dots), which still shows significant estimation error.

\textbf{Trinomial loss ablation.} Fig.~\ref{fig:vig_conv} compares the estimation error for two identical \sysname{} systems except that one did not receive trinomial loss training. Both use the measurement model from Eq.~\ref{eq:gauss}. The particle filter converges more quickly (74 filter updates compared to 146) and has lower average estimation error (314 m vs. 776 m) with trinomial loss. The additional training with the trinomial loss improves image retrieval performance with semi-positive image pairs and contributes towards improved particle filter performance.

\textbf{Multiple path result summary.} Table \ref{tab:summary_results} summarizes results comparing \sysname{} to the baseline on three test paths in Chicago. The first, C-1, is the path discussed previously (Fig.~\ref{fig:path}). C-2 is a 36~km long path of 104 time steps in southern Chicago and C-3 is a 41~km long path of 143 time steps in eastern Chicago. Our results demonstrate that \sysname{} consistently outperforms the baseline in final estimation error, average estimation error, and convergence time. The baseline did not converge on any of the paths tested.

\begin{table}[h]
\caption{Comparison of Results on Chicago Test Paths--\\ Baseline: Without trinomial loss and with exponential measurement model}
\vspace*{-.2in}
\label{tab:summary_results}
\begin{center}
\begin{tabular}{|cl|c|c|c|}
\hline
\multicolumn{2}{|c|}{\bf Metric and System Type}  & \bf C-1  & \bf C-2 & \bf C-3 \\\hline
\multirow{2}{*}{Average Error (m)} & Baseline & 1529 & 1186 & 2290\\
& \sysname{} & 314 & 948 & 2130\\\hline
\multirow{2}{*}{Final Error (m)} & Baseline & 1236 & 1175 & 1880\\
& \sysname{} & 21 & 11 & 53\\\hline
\multirow{2}{*}{Convergence Time (time steps)} & Baseline & - & - & -\\
& \sysname{} & 74 & 98 & 101\\\hline
\end{tabular}
\end{center} 
%
\caption{Comparison of Results on Singapore Test Paths}
\vspace*{-.2in}
\label{tab:sing_acc}
\begin{center}
\begin{tabular}{|cl|c|}
\hline
\multicolumn{2}{|c|}{\bf Metric and System Type}  & \bf Singapore  \\\hline
\multirow{2}{*}{Average Error (m)} & Baseline (CVM-Net \cite{Hu}) & 16.39 \\
& \sysname{} & 5.92 \\\hline
\end{tabular}
\end{center} \vspace*{-0.2in}
\end{table}

\subsection{Small-scale Test: Singapore}
\begin{figure}[t!]
\centering
  \includegraphics[height=0.8\linewidth]{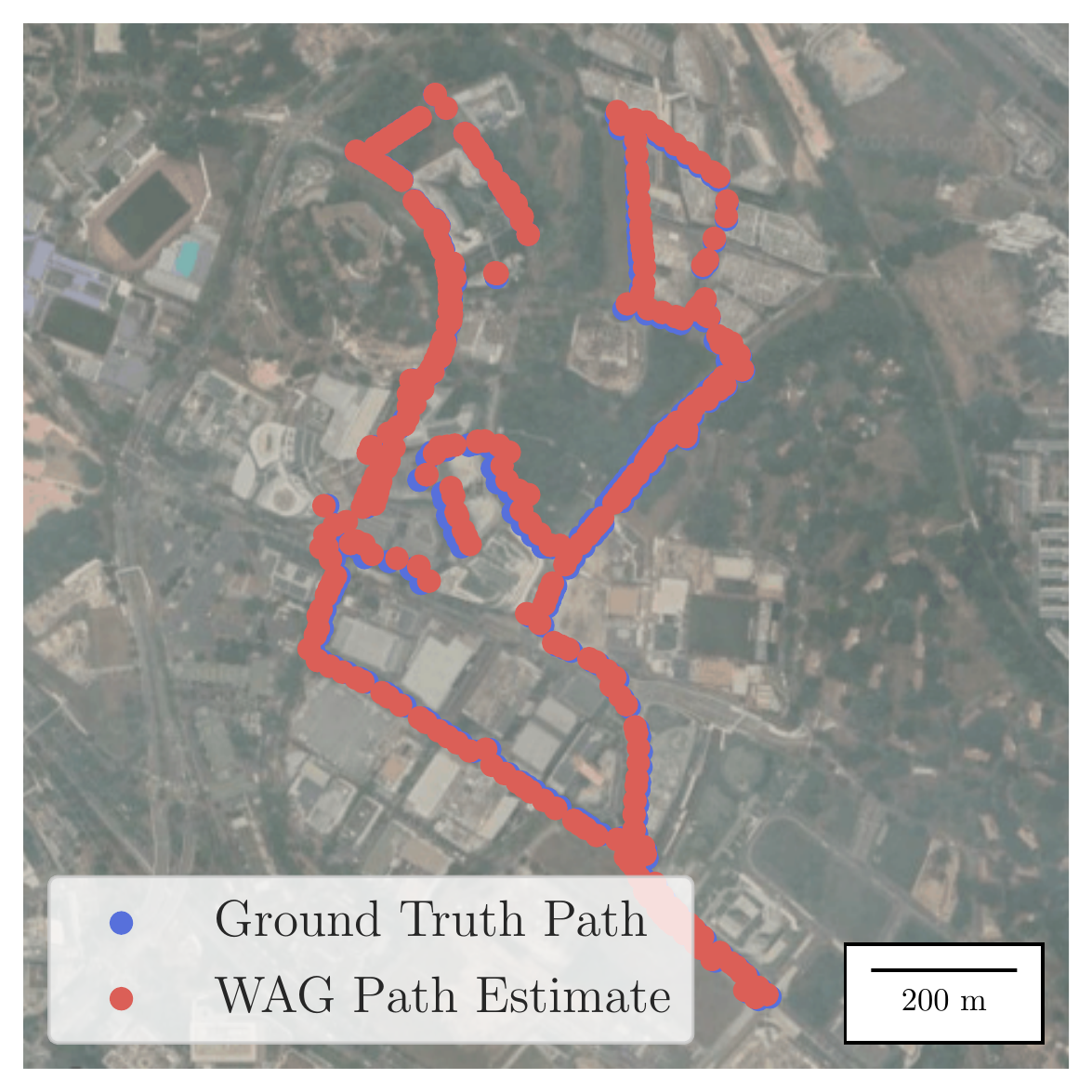}
  \caption{The ground-truth path in the One North area of Singapore and \sysname{}'s particle filter path estimate, which tracks the ground-truth closely.}
  \label{fig:sing_path} \vspace*{-0.2in}
\end{figure}

To demonstrate the ability of \sysname{} to perform well on smaller scale localization tasks like those demonstrated in \cite{Hu,Kim,ZhuVIGOR}, we have attempted to simulate the experiment performed by \cite{Hu}, which was done in a region of Singapore called ``One North''. We have produced a path similar to that in \cite{Hu} with ground images from Google Static API; the locations of these images are shown in Fig.~\ref{fig:sing_path} with \sysname{}'s estimate overlaid. The path is approximately 5 km long and the overhead satellite imagery that bounds it covers approximately 1.5 km$\times$1.5 km. 
The initial location of the agent is provided exactly as in \cite{Hu}, so this scenario can be viewed as an odometry drift reduction test. We did not have access to the neural network, CVM-Net, or the particle filter implementation used in \cite{Hu}, so we can only compare to their reported results. Ref.~\cite{Hu} uses visual odometry, but here we add 5\% noise to our simulated odometry, which is more than the average noise reported in \cite{Liu}, the source of the visual odometry algorithm used in \cite{Hu}. The results in Table \ref{tab:sing_acc} show that \sysname{} decreases the final estimation error by 63.9\%. 

\section{Conclusion}
\sysname{} presents a significant contribution towards the application of geolocalization systems to mobile robotics. Previous works have pushed the field further towards improving computer vision benchmarks, but have largely treated robotic applications as an afterthought. This work designs a system that improves aspects key to robotics-- in the coarser satellite image database that reduces computation and storage requirements and in the Gaussian measurement model that enables faster and more accurate particle filter convergence. 

\sysname{} accurately localizes across several hundred square kilometers of Chicago, a scale not previously demonstrated. \sysname{} has lower average error than the baseline as it localizes in this Chicago test area and it converges to location estimates faster. In addition to its impressive localization at a large scale, \sysname{} also outperforms \cite{Hu}'s reported results from localizing on a smaller scale of several dozen square kilometers in Singapore. Concurrently, \sysname{} requires $\approx$175 less satellite imagery and hence computes similarity matrices for particle filter measurement updates much faster than the baseline and requires many times less storage space for satellite embeddings. 

Overall, \sysname{} represents an improvement in the flexibility of geolocalization systems for mobile robotics. It maintains similar levels of accuracy to previous systems while reducing storage requirements, reducing computation, and enabling accurate localization on a city-wide scale as well as on a neighborhood-wide scale. Future work on this topic includes further optimization of computation to push the system closer to real-time, testing on real platforms in addition to in simulation, and improving localization accuracy beyond the scale of the satellite image tile size.

\bibliographystyle{IEEEtran}
\bibliography{root}

\newcommand{\noop}[1]{}
\begin{thebibliography}{10}
\providecommand{\url}[1]{#1}
\csname url@rmstyle\endcsname
\providecommand{\newblock}{\relax}
\providecommand{\bibinfo}[2]{#2}
\providecommand\BIBentrySTDinterwordspacing{\spaceskip=0pt\relax}
\providecommand\BIBentryALTinterwordstretchfactor{4}
\providecommand\BIBentryALTinterwordspacing{\spaceskip=\fontdimen2\font plus
\BIBentryALTinterwordstretchfactor\fontdimen3\font minus
  \fontdimen4\font\relax}
\providecommand\BIBforeignlanguage[2]{{%
\expandafter\ifx\csname l@#1\endcsname\relax
\typeout{** WARNING: IEEEtran.bst: No hyphenation pattern has been}%
\typeout{** loaded for the language `#1'. Using the pattern for}%
\typeout{** the default language instead.}%
\else
\language=\csname l@#1\endcsname
\fi
#2}}

\bibitem{Cai}
S.~Cai, Y.~Guo, S.~Khan, J.~Hu, and G.~Wen, ``Ground-to-aerial image
  geo-localization with a hard exemplar reweighting triplet loss,'' in
  \emph{2019 IEEE/CVF International Conference on Computer Vision (ICCV)},
  2019, pp. 8390--8399.

\bibitem{Tian}
Y.~Tian, C.~Chen, and M.~Shah, ``Cross-view image matching for geo-localization
  in urban environments,'' in \emph{IEEE Conference on Computer Vision and
  Pattern Recognition (CVPR)}, 2017, pp. 1998--2006.

\bibitem{Hu}
\BIBentryALTinterwordspacing
S.~Hu and G.~H. Lee, ``Image-based geo-localization using satellite imagery,''
  \emph{International J.\ of Computer Vision}, pp. 1205--1219, 2020. [Online].
  Available: \url{https://doi.org/10.1007/s11263-019-01186-0}
\BIBentrySTDinterwordspacing

\bibitem{Kim}
D.-K. Kim and M.~R. Walter, ``Satellite image-based localization via learned
  embeddings,'' \emph{2017 IEEE International Conference on Robotics and
  Automation (ICRA)}, pp. 2073--2080, 2017.

\bibitem{Shi}
Y.~Shi, L.~Liu, X.~Yu, and H.~Li, ``Spatial-aware feature aggregation for image
  based cross-view geo-localization,'' in \emph{NeurIPS}, 2019.

\bibitem{Liu2019}
L.~Liu and H.~Li, ``Lending orientation to neural networks for cross-view
  geo-localization,'' \emph{2019 IEEE/CVF Conference on Computer Vision and
  Pattern Recognition (CVPR)}, pp. 5617--5626, 2019.

\bibitem{Zhu2021}
S.~Zhu, T.~Yang, and C.~Chen, ``Revisiting street-to-aerial view image
  geo-localization and orientation estimation,'' in \emph{2021 Winter Conf.\ on
  Applications of Computer Vision}, 01 2021, pp. 756--765.

\bibitem{ZhuVIGOR}
------, ``Vigor: Cross-view image geo-localization beyond one-to-one
  retrieval,'' \emph{2021 IEEE/CVF Conference on Computer Vision and Pattern
  Recognition (CVPR)}, pp. 5316--5325, 2021.

\bibitem{Xia}
Z.~Xia, O.~Booij, M.~Manfredi, and J.~F.~P. Kooij, ``Cross-view matching for
  vehicle localization by learning geographically local representations,''
  \emph{IEEE Robotics and Automation Letters}, vol.~6, pp. 5921--5928, 2021.

\bibitem{Viswanathan}
A.~Viswanathan, B.~R. Pires, and D.~F. Huber, ``Vision based robot localization
  by ground to satellite matching in gps-denied situations,'' \emph{2014
  IEEE/RSJ International Conference on Intelligent Robots and Systems}, pp.
  192--198, 2014.

\bibitem{viswanathan2016}
A.~Viswanathan, B.~R. Pires, and D.~Huber, ``Vision-based robot localization
  across seasons and in remote locations,'' in \emph{2016 IEEE International
  Conference on Robotics and Automation (ICRA)}, 2016, pp. 4815--4821.

\bibitem{jacobs}
N.~Jacobs, S.~Satkin, N.~Roman, R.~Speyer, and R.~Pless, ``Geolocating static
  cameras,'' in \emph{2007 IEEE 11th International Conference on Computer
  Vision}, 2007, pp. 1--6.

\bibitem{bansal}
\BIBentryALTinterwordspacing
M.~Bansal, H.~S. Sawhney, H.~Cheng, and K.~Daniilidis, ``Geo-localization of
  street views with aerial image databases,'' in \emph{Proceedings of the 19th
  ACM International Conference on Multimedia}, ser. MM '11.\hskip 1em plus
  0.5em minus 0.4em\relax New York, NY, USA: Association for Computing
  Machinery, 2011, p. 1125–1128. [Online]. Available:
  \url{https://doi-org.libproxy.mit.edu/10.1145/2072298.2071954}
\BIBentrySTDinterwordspacing

\bibitem{Workman}
S.~Workman, R.~Souvenir, and N.~Jacobs, ``Wide-area image geolocalization with
  aerial reference imagery,'' \emph{2015 IEEE International Conference on
  Computer Vision (ICCV)}, pp. 3961--3969, 2015.

\bibitem{bromley}
J.~Bromley, J.~Bentz, L.~Bottou, I.~Guyon, Y.~Lecun, C.~Moore, E.~Sackinger,
  and R.~Shah, ``Signature verification using a "siamese" time delay neural
  network,'' \emph{International Journal of Pattern Recognition and Artificial
  Intelligence}, vol.~7, p.~25, 08 1993.

\bibitem{Vo}
N.~N. Vo and J.~Hays, ``Localizing and orienting street views using overhead
  imagery,'' in \emph{ECCV}, 2016.

\bibitem{Rodrigues}
\BIBentryALTinterwordspacing
R.~Rodrigues and M.~Tani, ``Are these from the same place? seeing the unseen in
  cross-view image geo-localization,'' in \emph{2021 IEEE Winter Conference on
  Applications of Computer Vision (WACV)}.\hskip 1em plus 0.5em minus
  0.4em\relax Los Alamitos, CA, USA: IEEE Computer Society, jan 2021, pp.
  3752--3760. [Online]. Available:
  \url{https://doi.ieeecomputersociety.org/10.1109/WACV48630.2021.00380}
\BIBentrySTDinterwordspacing

\bibitem{Cao}
R.~Cao, J.~Zhu, Q.~Li, Q.~Zhang, Q.~Li, B.~Liu, and G.~Qiu, ``Learning
  spatial-aware cross-view embeddings for ground-to-aerial geolocalization,''
  in \emph{ICIG}, 2019.

\bibitem{lin}
T.-Y. Lin, S.~Belongie, and J.~Hays, ``Cross-view image geolocalization,'' in
  \emph{2013 IEEE Conference on Computer Vision and Pattern Recognition}, 2013,
  pp. 891--898.

\bibitem{lin2015}
T.-Y. Lin, Y.~Cui, S.~Belongie, and J.~Hays, ``Learning deep representations
  for ground-to-aerial geolocalization,'' in \emph{2015 IEEE Conference on
  Computer Vision and Pattern Recognition (CVPR)}, 2015, pp. 5007--5015.

\bibitem{crisan}
D.~Crisan and A.~Doucet, ``A survey of convergence results on particle
  filtering methods for practitioners,'' \emph{IEEE Transactions on signal
  processing}, vol.~50, no.~3, pp. 736--746, 2002.

\bibitem{Liu}
P.~Liu, M.~Geppert, L.~Heng, T.~Sattler, A.~Geiger, and M.~Pollefeys, ``Towards
  robust visual odometry with a multi-camera system,'' in \emph{2018 IEEE/RSJ
  International Conference on Intelligent Robots and Systems (IROS)}, 2018, pp.
  1154--1161.

\end{thebibliography}

\end{document}